\documentclass[runningheads]{llncs}

\usepackage{eccv}

\usepackage{eccvabbrv}

\usepackage{graphicx}
\usepackage{booktabs}
\usepackage{multirow}
\usepackage{arydshln}
\usepackage{subcaption}
\usepackage{wrapfig}
\usepackage[symbol]{footmisc}
\usepackage[ruled,linesnumbered]{algorithm2e}

\usepackage[accsupp]{axessibility}  

\usepackage[pagebackref,breaklinks,colorlinks,citecolor=eccvblue]{hyperref}

\usepackage{orcidlink}

\begin{document}

\title{Enhanced Sparsification via Stimulative Training} 


\author{Shengji Tang\thanks{These authors contributed equally.}\inst{1} \and Weihao Lin\textsuperscript{$\ast$}\inst{1} \and Hancheng Ye\inst{3} \and \\Peng Ye\inst{1} \and Chong Yu\inst{2} \and Baopu Li\inst{4} \and Tao Chen\thanks{Corresponding author}\inst{1}}

\authorrunning{Tang et al.}

\institute{School of Information
Science and Technology, Fudan University, Shanghai, China \and
Academy for Engineering and Technology, Fudan University, Shanghai, China \and Shanghai AI Laboratory, Shanghai, China \and Independent Researcher \\
\email{eetchen@fudan.edu.cn}}

\maketitle

\begin{abstract}
Sparsification-based pruning has been an important category in model compression. Existing methods commonly set sparsity-inducing penalty terms to suppress the importance of dropped weights, which is regarded as the suppressed sparsification paradigm. However, this paradigm inactivates the dropped parts of networks causing capacity damage before pruning, thereby leading to performance degradation. To alleviate this issue, we first study and reveal the relative sparsity effect in emerging stimulative training and then propose a structured pruning framework, named STP, based on an enhanced sparsification paradigm which maintains the magnitude of dropped weights and enhances the expressivity of kept weights by self-distillation. Besides, to find an optimal architecture for the pruned network, we propose a multi-dimension architecture space and a knowledge distillation-guided exploration strategy. To reduce the huge capacity gap of distillation, we propose a subnet mutating expansion technique. Extensive experiments on various benchmarks indicate the effectiveness of STP. Specifically, without fine-tuning, our method consistently achieves superior performance at different budgets, especially under extremely aggressive pruning scenarios, e.g., remaining 95.11\% Top-1 accuracy (72.43\% in 76.15\%) while reducing 85\% FLOPs for ResNet-50 on ImageNet. Codes will be released soon.   
\end{abstract}

\section{Introduction}
\label{sec:intro}
Recently, the model pruning~\cite{blalock2020state,liu2017learning,han2015deep} technique, which removes redundant parameters to obtain a compact network from an over-parameterized one, has become a mainstream and general model compression method. Early pruning methods~\cite{lecun1989optimal,hassibi1992second,han2015learning} focus on designing delicate metrics, e.g., $L_1$ score~\cite{han2015learning} or feature activation~\cite{hu2016network}, to measure the importance of parameters and discard the unimportant ones. However, directly removing parameters from the original network often causes dramatic performance degradation, especially in structured pruning where groups of filters are removed. To alleviate the performance degradation, emerging works~\cite{yang2019deephoyer,he2017channel,ding2018auto,wang2019structured,wang2021neural} are dedicated to sparsifying parameters before pruning or sparsifying and pruning alternately. The sparsification aims to alter the distribution of parameters and transfer the network expressivity to the kept parameters, which helps to maintain performance after pruning. In current literature, the penalty terms~\cite{han2015learning,wen2016learning,bai2021dual} are commonly employed as regularization to introduce sparsification. They suppress the importance of dropped parameters, e.g., encouraging the pruned parameters to converge to zero~\cite{louizos2018learning}, which is regarded as the suppressed sparsification paradigm in this manuscript. However, it is pointed out that the suppressed sparsification paradigm commonly causes expressivity damage of the unpruned network during the sparsification procedure~\cite{ding2018auto,wang2021neural}, which limits the pruning upper bound. Numerous works have attempted to mitigate it by using a small regime of penalty stength~\cite{liu2017learning,wen2016learning} or well-designed penalty coefficients~\cite{wang2019structured,wang2021neural,ding2018auto}, but only alleviates rather than completely solves this issue. The reason is that the suppressed sparsification paradigm introduces explicit regularization to forcibly drive the pruned parts toward zero and obtain a sparse distribution that is easier to prune. \textbf{This process is nearly equivalent to removing the corresponding parameters, causing structural damage and loss of network capacity.} It is detrimental to the preservation of expressive capability before pruning. 
\begin{figure}[t]
\centering
\begin{minipage}[t]{0.49\linewidth}
\includegraphics[width=\linewidth]{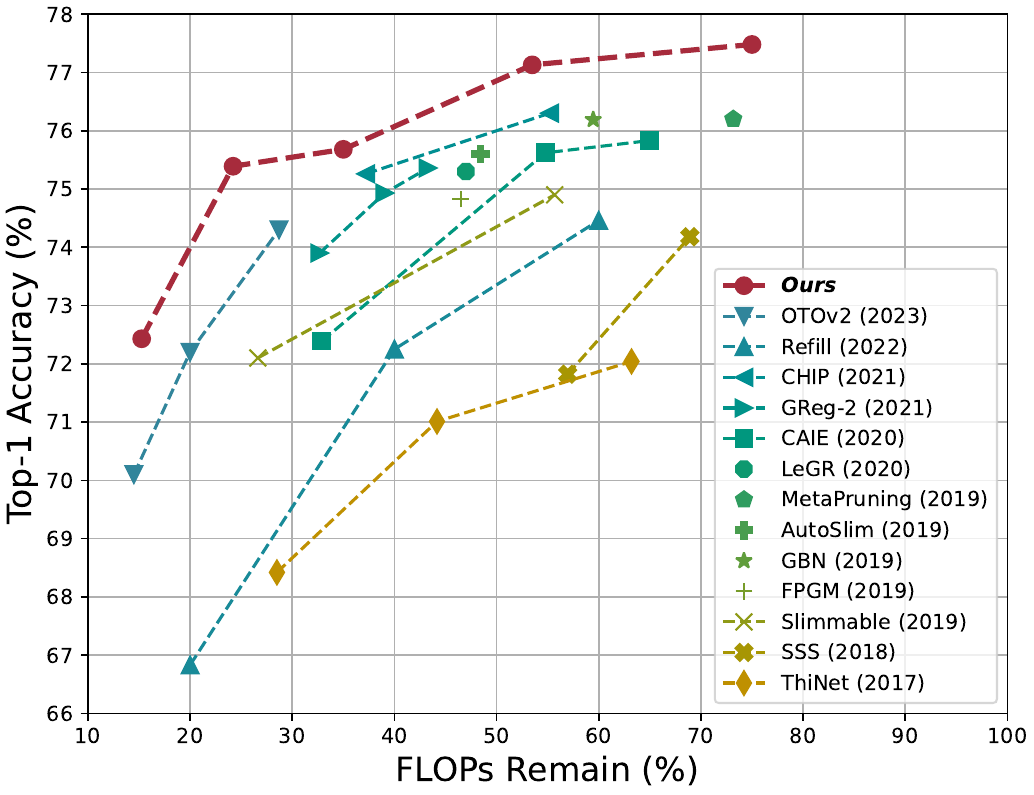}
\caption{ResNet-50 on ImageNet dataset. Top-1 accuracy (\%) and remaining FLOPs (in percentage) are reported. Without fine-tuning, our method can still obtain an optimal Pareto frontier of efficiency and performance compared with other methods.} 
\label{fig:r50-img}
\end{minipage}
\hfill
\begin{minipage}[t]{0.49\linewidth}
\includegraphics[width=\linewidth]{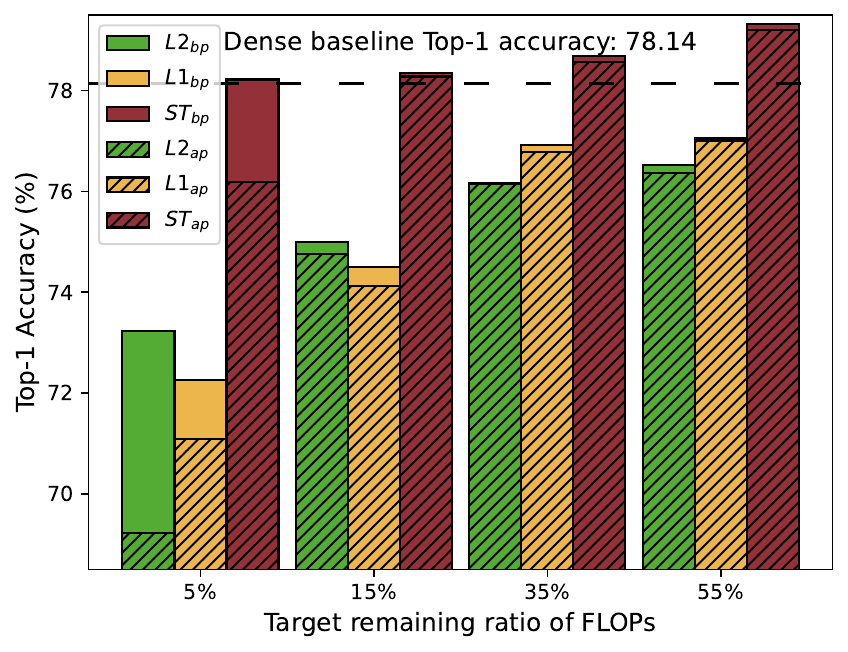}
\caption{Different sparsification methods, namely $L_1$, $L_2$, and ST, on CIFAR-100 dataset under different FLOPs. The subscripts \textbf{ap} and \textbf{bp} represent the performance of networks \textbf{after} and \textbf{before} pruning, respectively.
}
\label{fig:sp_method_cmp}
\end{minipage}
\vspace{-15pt}
\end{figure}
\par
\begin{figure*}[t]
  \centering
  \includegraphics[width=\linewidth]{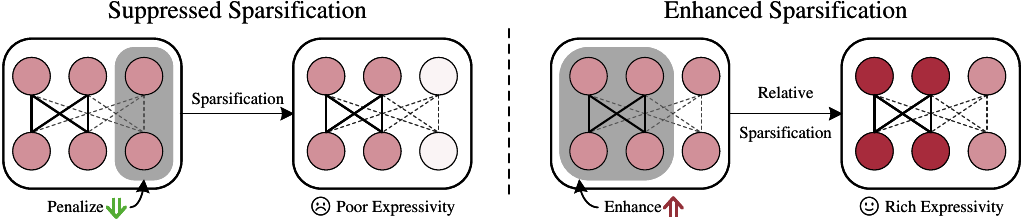}
  \caption{Comparison of suppressed and enhanced sparsification. The exemplified fully connected layer has three input neurons and three output ones. The solid black lines denote the parameters to remain, while the dashed ones are pruned. The parameters connected to a redder neuron have larger magnitudes.
  }
  \label{fig:suppress_vs_enhance}
  \vspace{-10pt}
\end{figure*}

 From the dual respect of suppression, we consider achieving relative sparsity by enhancing the kept parts while maintaining the magnitude of the to-be-pruned parts, termed enhanced sparsification paradigm. Fig.~\ref{fig:suppress_vs_enhance} provides an intuitive comparison of the suppressed and enhanced sparsification. The core consideration behind the enhanced sparsification paradigm is that the enhanced kept parts can occupy a greater significance in the contribution to the overall network's expressive capability while avoiding damage to the entire network. It is noteworthy that relative sparsity differs from the commonly used definition of sparsity. Relative sparsity refers to a situation where some parameters in the network have a notably greater ratio (not the absolute value) than others on certain metrics, such as the $L_1$ norm. To realize the enhanced sparsification paradigm, we focus on the original purpose of sparsification, which transfers the expressivity from the whole to the kept part for lossless pruning. A popular technique for transferring is knowledge distillation (KD)~\cite{hinton2015distilling,kim2021self}, which aligns the output logits or features to transfer the dark knowledge. Recently, Stimulative Training (ST)~\cite{ye2022stimulative,ye2023stimulative} utilizes self-distillation to transfer the knowledge from the main network to its weight-sharing subnets and benefits both the main network and subnets. We consider the self-distillation process in ST can be seen as a natural enhancement. As shown in Fig.~\ref{fig:weight_distribution}, we study the parameter distribution in ST and observe that without an explicit sparsity constraint, ST results in prominent relative sparsity, which will be further discussed in Sec.~\ref{sec:sparsity_effect}. 
\par Based on the relative sparsity effect in ST, we propose a new enhanced sparsification paradigm and build a one-stage multi-dimension framework for structured pruning, namely ST guided pruning (STP). The core of the method lies in, given a subnet, utilizing the self-distillation in ST to distill the main network to the weight-sharing subnet. This enhances the importance of the subnet, achieves relative sparsification, and thereby facilitates lossless structured pruning.\footnote{It is noted that since unstructured pruning hardly promises practical speedup, STP only focuses on structured pruning.} Besides, STP introduces three key designs to guarantee a high-performance compact network: 1) KD guided exploration. It is a heuristic searching strategy driven by the KD loss to obtain an outstanding subnet for sparsification. We introduce KD guided exploration to gradually explore surpassing subnet architectures during ST; 2) Multi-dimension sampling. The original ST~\cite{ye2022stimulative} only considers sampling subnets with different depths (total layer numbers). To formulate a more general pruning framework and chase for a better trade-off between sparsity and performance, we extend the sampling dimension of ST to both depth and width (the output channel number of each layer) and design a multi-dimension FLOPs-aware sampling space to fit the desired FLOPs target correctly; 3) Subnet mutating expansion. The FLOPs-aware sampling space leads to monotonic subnet capacity, which harms the performance of the main network~\cite{tang2023boosting}. The subnet mutating expansion generates larger subnets to enrich the subnet capacity, improving the main network.  
\par We summarize our contributions as follows: 1) We first reveal the relative sparsity effect of ST, which renders ST to be an enhanced sparsification paradigm maximumly retaining the capacity of the unpruned network, thus alleviating performance drop after pruning; 2) Based on the enhanced sparsification paradigm, we propose a pruning framework, named stimulative training guided pruning (STP). The STP includes a KD guided architecture exploration method to find optimal subnet architectures, a multi-dimension sampling to better trade-off between sparsity and performance, and a subnet mutating expansion scheme to introduce various support subnets for further performance improvement; 3) Extensive empirical results and comparisons on various mainstream models (ResNet-50, WRN28-10, MobileNetV3, ViT, Swin Transformer) and datasets (CIFAR-100, TinyImageNet, ImageNet, COCO) show that STP obtains compact networks with high performance and extremely low FLOPs without fine-tuning, e.g., preserving 95.11\% performance (72.43\% in 76.15\% Top-1 accuracy) while reducing 85\% FLOPs on ImageNet (as shown in Fig.~\ref{fig:r50-img}).

\section{Related Work}

\subsection{Sparsification-based Pruning}
Existing sparsification-based pruning methods can be divided into two categories: static sparsification and dynamic sparsification. Static sparsification aims at imposing a global and unchanged penalty strength to all parameters of the model~\cite{liu2017learning,learning_numbers,lin2019toward,liu2015sparse,wen2016learning,bai2021dual,han2015learning, louizos2018learning,ding2021resrep,xia2023structured}. The penalty terms in these works are designed to be a norm-based regularization item w.r.t. model parameters. 
Different from static sparsification methods, dynamic sparsification considers the individual and time-varying sparsification strength of different parameters~\cite{wang2019structured,wang2021neural,ding2018auto,wang2019structured, ye2022beta, ye2023performance, yu2022unified, chen2021chasing, bai2021dual}. AFP~\cite{ding2018auto} utilizes auto-balanced regularization for different weights to transfer the representational capacity from the whole network to the kept part. Greg~\cite{wang2021neural} merely applies sparsification on the dropped parameters and grows the penalty term large gradually to achieve higher sparsity. 
Previous works mostly follow the suppressed sparsification paradigm, which imposes constraints on dropped parameters, causing expressivity damage before pruning. Differently, our method achieves relative sparsity by enhancing kept parameters and maintaining the magnitude of dropped ones via stimulative training.

\subsection{Knowledge Distillation in Pruning}
In general, pruning a network inevitably deteriorates its performance. To compensate for the performance degradation, conducting knowledge distillation (KD) after pruing has been proved feasible~\cite{aflalo2020knapsack,aghli2021combining,xia2022structured,ma2021good,neill2021deep,chen2021knowledge,li2020few}. In~\cite{aghli2021combining}, the authors firstly obtain a pruned network via APoZ~\cite{hu2016network} and then transfer knowledge from the unpruned network to the pruned one via a cosine-similarity-based loss. The authors in~\cite{ma2021good} apply a Kullback-Leibler-divergence-based loss for KD and verify its effectiveness in both one-shot and iterative magnitude pruning. Built upon iterative magnitude pruning, the authors in~\cite{neill2021deep} propose a cross-correlation-based KD loss to better align the pruned network with the unpruned one. The aforementioned methods treat KD as a performance booster isolated from pruning. Differently, we treat KD itself as a pruning method, not merely a support technique. Our proposed method incorporates KD into pruning, utilizing self-distillation for sparsification and KD loss to obtain the pruning mask. 

\subsection{Stimulative Training}
Stimulative training (ST)~\cite{ye2022stimulative} aims to boost the main network by transferring the knowledge from the main network to each depth-wise subnet. In ST, the main network can be seen as an ensemble of shallow subnets, and the final performance is determined by each subnet under the unraveled view~\cite{veit2016residual,sun2022low}. Thus, while training the main network, ST randomly skips some layers to generate subnets and utilizes knowledge distillation to supervise each subnet. The experiments in ST show that ST can obtain performance improvement on the main network while bringing various subnets with marginal performance reduction compared with the main network. This additional benefit of subnets suggests that ST can achieve simultaneous enhancement at both local and global of the main network, which inspires us to study the distribution of parameters in ST and exploit ST as an enhanced sparsification paradigm for pruning. 

\section{Method}

\subsection{Problem Formutation}
\label{sec:problem}
Pruning aims to eliminate parameters from a given network $\mathcal{N}$ with parameters $\theta$ to obtain a high-performance compact subnet $\mathcal{N}_s$ with retaining parameters $\theta_s$ with specific resource constraint. Although $\mathcal{N}$ can be given with pretrained weights, our method concentrates on pruning from randomly initialized parameters, same as~\cite{bai2021dual,hou2022chex}. The pruning process can be abstracted as a function $F(\cdot)$, which takes $\theta$ as input and output $\theta_s$, $\theta_s=F(\theta)$\footnote{Because the dependence of dataset is upon the pruning algorithm, for simplicity, we omit the given dataset $D$ as input.}. General pruning mainly consists of two categories of transformations 1) $\mathcal{M} = F_1(\theta)$, which generates a binary mask to determine remaining and dropping parts 2) $\hat{\theta} = F_2(\theta)$, which adjusts the parameter distribution for eliminating with low damage. A typical pruning process can be denoted as: 
\begin{align}
    \theta_s = \mathcal{M} \odot \hat \theta = F_1(\theta) \odot F_2(\theta).
\end{align}
The exact format depends on the specific pruning algorithm. For one-shot pruning, $F_1$ is applied once~\cite{KDSG17} while alternately for iterative pruning~\cite{frankle2018lottery}. We denote the dropped parameters as $\theta_d = \theta \setminus \theta_s$. The specification process belongs to $F_2(\cdot)$ and a typical method~\cite{wang2019structured,wang2021neural,bai2021dual} introduces $L_2$ norm penalty term for $\theta_d$, which optimizes the following objective:
\begin{align}
    \mathcal{L}_{total}(\theta;D) = \mathcal{L}_{task}(\theta;D) +  \frac{\lambda}{2}||\theta_d||_2^2,
\end{align}
where $\mathcal{L}_{task}$ is the task loss; $D$ is the given dataset;$\lambda$ is the coefficient of penalty term. The penalty term drives $||\theta_d||$ to converge to zero and eliminates the importance of $\theta_d$ for painless removal, which belongs to suppressed paradigm. Whereas $\theta_d$ is a part of $\theta$, suppression of $\theta_d$ reduces the capacity of $\theta$, which limits the upper bound of $\theta_s$ and compromises the final performance. 

\subsection{Relative Sparsity Effect in Stimulative Training}
\label{sec:sparsity_effect}
To boost the performance of the main network, stimulative training (ST) views a network as an ensemble of multiple shallow subnets and transfers the knowledge from the main network to each weight-shared subnet. Formally, ST optimizes the following loss:
\begin{align}
\label{formulation:ST}
\mathcal{L}_{ST} = CE(p(x;\theta_m), y) + \lambda KL(p(x;\theta_m)|| p(x;\theta_s)),
\end{align}
\begin{figure}[t]
\centering 

\begin{subcaptionblock}[t]{0.24\linewidth}
\centering
\label{Fig.sub.1}
\includegraphics[width=\linewidth]{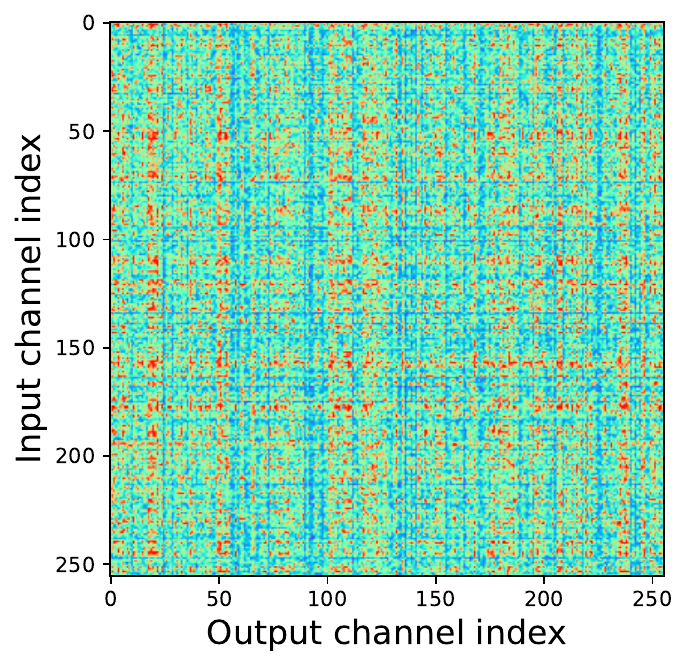}
\caption{Baseline, layer 3-1}
\end{subcaptionblock} 
\begin{subcaptionblock}[t]{0.24\linewidth}
\centering
\label{Fig.sub.2}
\includegraphics[width=\linewidth]{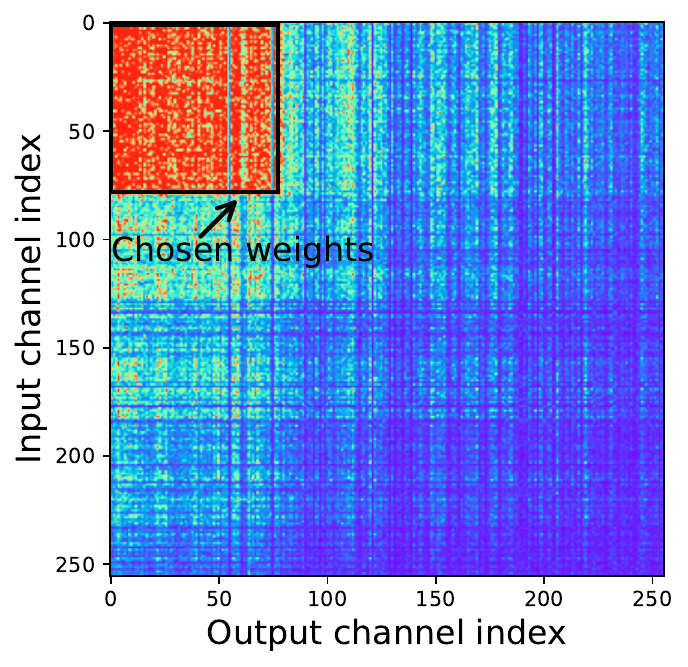}
\caption{ST, layer 3-1, 0.3}
\end{subcaptionblock}
\begin{subcaptionblock}[t]{0.24\linewidth}
\centering
\label{Fig.sub.3}
\includegraphics[width=\linewidth]{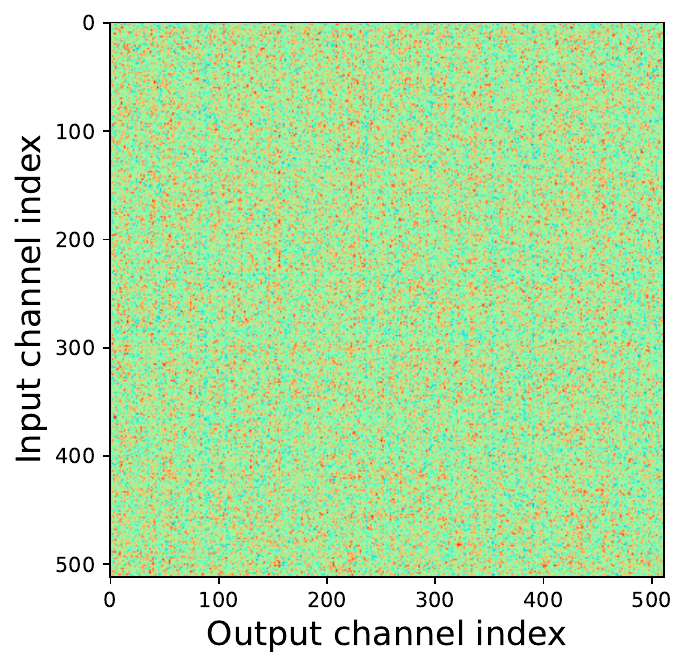}
\caption{Baseline, layer 4-2}
\end{subcaptionblock}
\begin{subcaptionblock}[t]{0.24\linewidth}
\centering
\label{Fig.sub.4}
\includegraphics[width=\linewidth]{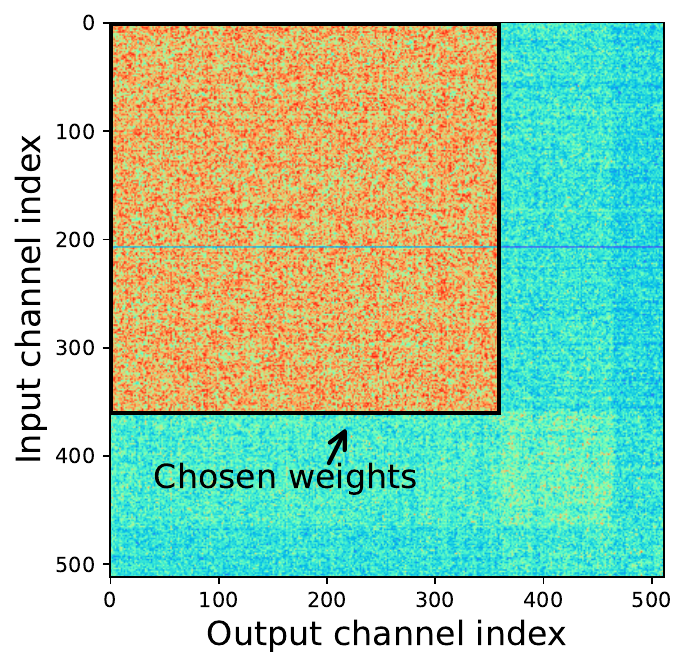}
\caption{ST, layer 4-2, 0.7}
\end{subcaptionblock}

\caption{The convolutional weight magnitude distribution of baseline (standard training without sparsification) and stimulative training (ST)~\cite{ye2022stimulative} with different sparsity ratios (0.3 and 0.7) in different layers (3-1 and 4-2) on ResNet-50, where ``3-1” and ``4-2” are the layer index, e.g., given ResNet-50 containing four stages with layers [3,4,6,3], ``3-1” represents the first layer of the third stage. The color represents the \textbf{relative} magnitude comparison. The redder regions indicate the relatively larger magnitudes, and vice versa. The regions encircled by the black rectangle represent the chosen (kept) weights. Given a structured architecture, ST exhibits a significant relative sparsity effect, enhancing chosen weights accurately.}
\label{fig:weight_distribution}
\vspace{-10pt}
\end{figure}
where $x$ and $y$ are the input images and ground truth labels, respectively; $CE(\cdot)$ is the cross entropy; $KL(\cdot)$ is the Kullback-Leibler divergence; $\lambda$ is the balance coefficient; $p(x;\theta)$ is the output probability with $x$ and $\theta$; $\theta_m$ and $\theta_s$ are the parameters of the main network and subnet, respectively, and $\theta_s$ is the subset of $\theta_m$. According to the optimization of the second term of $\mathcal{L}_{ST}$, \ie, $KL(p(x;\theta_m)||p(x;\theta_s))$, ST can gradually transfer the representative ability from the whole parameters ($\theta_m$) to the partial ones ($\theta_s$), causing the enhancement of subnet weights and implying the potential of relative sparsity. 
\par To investigate the relative sparsity effect in ST, we conduct ST by fixing a specific subnet on ResNet-50 and visually show the magnitude of a convolutional layer's weights. Notely, while the original ST only samples subnet in the depth dimension, we extend it into the width dimension. As shown in Fig.~\ref{fig:weight_distribution}, compared with the random weights' distribution in the baseline, \ie, standard training with stochastic gradient descent, ST can significantly enhance the chosen weights, causing the weights' magnitude relatively concentrating on the chosen weights (the black rectangle). It is worth noting that Fig.~\ref{fig:weight_distribution} reports the relative comparison, see Fig.~\ref{fig:width_distribution} for detailed absolute values. To further display the effect of different sparsification methods before and after pruning, we conduct experiments with different target sparsity. As shown in Fig.~\ref{fig:sp_method_cmp}, the $L$ norm sparsifications harm the performance before pruning (before removing the dropped parameters), while ST maintains and even improves the performance. It suggests ST introduces a moderate and relative sparsity effect via KD to keep the capacity of the dropped parts, which encourages learning better sparsity distribution.

\subsection{Framework}
\begin{figure*}[t]
  \centering
  \includegraphics[width=\linewidth]{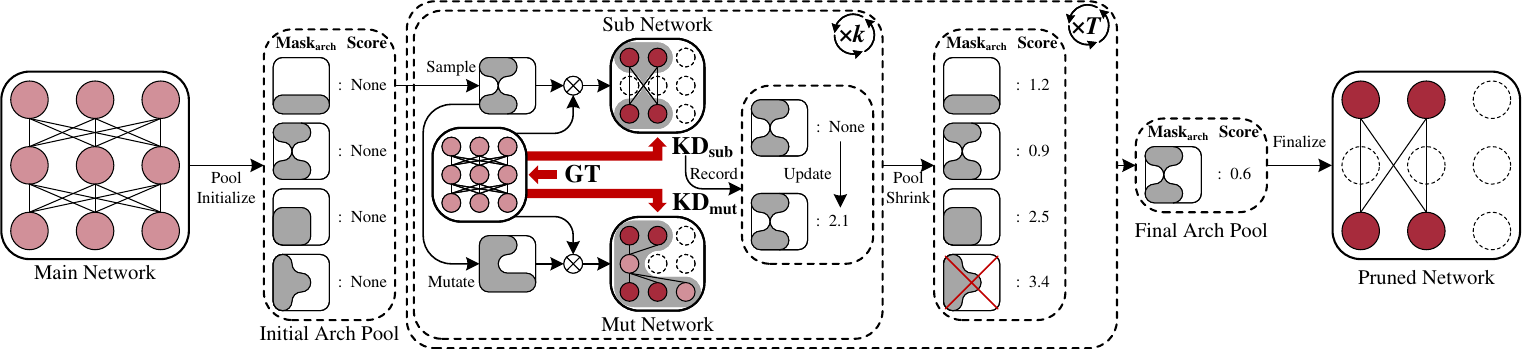}
  \caption{The framework of the proposed stimulative training guided pruning (STP). The recursive arrows mean repeating the corresponding process multiple times; bold red arrows represent computing loss and supervising. STP gradually explores the optimal subnet architecture guided by the KD loss and utilizes stimulative training to enhance the subnet with the support of mutating networks. Finally, the subnet can be separated from the main network as a structured pruned network with low FLOPs. $k$ is a hyper-parameter, and a typical value is the iteration number of an epoch.}
  \label{fig:framework}
  \vspace{-12pt}
\end{figure*}

The proposed ST guided pruning (STP) framework is illustrated in Fig.\ref{fig:framework}. In detail, given a dense network, \ie, the main network and a target FLOPs constraint, we initialize an architecture pool containing structured subnet architectures sampled from multi-dimension FLOPs-aware sampling space. The training starts with randomly initialized parameters. In each training iteration, the procedure can be divided into five steps: 1) sample a subnet architecture based on the score in the architecture pool; 2) forward the main network to generate the main output, compute and backward the cross entropy between ground truth and main output; 3) apply the sampled subnet architecture mask into the main network to construct a weight-shared subnet and supervise it using the main output by knowledge distillation. 4) mutate the sampled architecture mask multi-dimensionally into a larger architecture mask, which is utilized to generate a weight-shared mutating network and supervise it using the main output; 5) remove the subnet architecture with the highest score (KD loss) in the architecture pool every $k$ iterations. After $T$ training iterations, only one subnet architecture mask exists in the architecture pool. We apply it to the main network to obtain the final pruned network while the other parameters are deserted. A typical value of $k$ is the iteration number of an epoch. The pseudo-code is shown in the Appendix. Three critical designs of STP are introduced as follows.

\noindent \textbf{Stimulative Training Sparsification and Multi-dimension Sampling.}
As stated in Sec.~\ref{sec:sparsity_effect}, inspired by the relative sparsity effect in ST, we exploited it for enhanced sparsification. However, the original ST sampling space only considers the depth dimension (all possible layer combinations), ignoring the more fine-grained width dimension (output channels). The relatively coarse-grained depth sampling space cannot fully explore the trade-off between sparsity and performance. Thus, different from the original ST, we build a multi-dimension sampling space, including both depth and width, to handle general pruning cases in a fine-grained manner, which provides more possibilities for finding a high-performance architecture. Exemplary sampled subnets are referred to in Appendix.

Besides the multi-dimension design, we impose a FLOPs-aware constraint on the sampling space, only sampling the subnets with target FLOPs. It can guarantee a more accurate practical speedup ratio compared with sparsity constraint. Because the sampled subnets generally possess much lower capacity than the main network, we adopt the normalized Kullback-Leibler divergence (KL-)~\cite{ye2023stimulative} to eliminate the capacity gap for higher performance.
\\
Formally, given a structured mask $\mathcal{M}_s$, the parameters of subnets are $\theta_s=\mathcal{M}_s \odot \theta_m$. ST sparsification is optimized by:
\begin{equation}
\label{formu:l_stc}
\begin{aligned}
\mathcal{L}_{STS} &= KL_-(\hat{p}(x;\theta_m)||\hat{p}(x;\theta_s)) = \sum_{i=1}^{N} \hat{p}_{i}(x;\theta_m) \log \frac{\hat{p}_{i}(x;\theta_m)}{\hat{p}_{i}(x;\theta_s)}.
\end{aligned}
\end{equation}
$\hat{p}(x;\theta)$ is the normalized probability which is donated as: 
\begin{align}
\hat{p}_i(x;\theta) = \frac {e^{\frac{Z_i(x;\theta)}{\|Z(x;\theta)\|}}} {\sum_{j=1}^{N} e^{\frac{Z_j(x;\theta)}{\|Z(x;\theta)\|}}},
\end{align}
where $Z(x;\theta)$ is the output logits. $L_{STS}$ is essentially a knowledge distillation loss, and illustrated as $KD_{sub}$ in Fig.\ref{fig:framework}. The subnets will also be called target subnets in the following section.
\\
\noindent \textbf{Subnet Mutating Expansion.}
To mitigate the performance degradation resulting from focusing on training tiny subnets~\cite{tang2023boosting}, we introduce additional larger subnets to provide support during training. As a distinction, we refer to them as support subnets. Considering that randomly sampling larger support subnets will disrupt the parameter magnitude concentration of target subnets, we propose to mutate and expand the width and depth of target subnets to generate support subnets.
\par Formally, given a target subnet with layer-wise sparsity ratio $S_{tar}=\{S_{tar}^1, S_{tar}^2\\, ...,S_{tar}^L\}$, for each $S_{tar}^i$ we uniformly sample a larger ratio $S_{sup}^i$ which satisfies:  
\begin{align}
\label{formu:sup_sample}
S_{sup}=\{S_{sup}^1, S_{sup}^2, ...,S_{sup}^L\},S_{sup}^i \sim U(S_{tar}^i, 1),
\end{align}
where $U(\cdot)$ denotes the uniform distribution. The support subnet will be supervised by the main network similar to Formulation \ref{formu:l_stc}, denoted as:
\begin{align}
\mathcal{L}_{SME} = KL_-(\hat{p}(x;\theta_m)||\hat{p}(x;\theta_{sup})),
\end{align}
where $\theta_{sup}$ is the parameters of the support subnet and $\mathcal{L}_{SME}$ is illustrated as $KD_{mut}$ in Fig.~\ref{fig:framework}.
Formulation~\ref{formu:sup_sample} can guarantee that the support subnet contains the whole parameters of the sampled target subnet. Thus, training the support subnet also enhances the target subnet. 

\noindent \textbf{Knowledge Distillation guided Exploration.}
To explore the superior architecture of target subnets, we construct a candidate pool containing numerous architectures satisfying the FLOPs target. KD loss ($L_{STS}$) is a reliable indicator to evaluate the generation and performance gap between subnets and the main network. Under the constraint of FLOPs, a lower KD loss implies a more superior architecture. Thus, we heuristically utilize KD loss to guide the gradual shrinking of the pool and obtain a high-performance architecture.   
\par Specifically, suppose there is architecture pool with $N$ candidate masks, denoted as $\mathcal{P}=\{\mathcal{M}_1, \mathcal{M}_2, ..., \mathcal{M}_N \}$. Each architecture has a corresponding score, denoted as $Sco=\{Sco_1, Sco_2, ..., Sco_N\}$. While training a target subnet with $\mathcal{M}_i$, we record the $\mathcal{L}_{STS}$ and update the corresponding score via exponential moving average (EMA)~\cite{he2020momentum}, which can be denoted as:
\begin{align}
Sco_i := (1-\alpha Sco_i) + \alpha \mathcal{L}_{STS},
\end{align}
where $\alpha \in (0,1)$ is the updating ratio. The pool is shrunk by removing the architectures with the highest scores every $k$ iterations. After $T$ rounds of shrinking, the pool contains a single architecture, which is chosen as the pruned network.
\begin{table*}[t]
\centering
\caption{Comparison of different pruning methods on CIFAR-100. We report the Top-1 accuracy(\%) of dense and pruned networks with different remaining FLOPs.}
\setlength\tabcolsep{10pt}
\resizebox{\linewidth}{!}{
\begin{tabular}{l|ccc|ccc|ccc}
\hline
\multirow{2}{*}{Method} & \multicolumn{3}{c|}{ResNet-50 (Acc: 78.14)}                       & \multicolumn{3}{c|}{MBV3 (Acc: 78.09)}                           & \multicolumn{3}{c}{WRN28-10 (Acc: 82.17)}                          \\ \cline{2-10} 
                        & \multicolumn{1}{c|}{15\%}  & \multicolumn{1}{c|}{35\%}  & 55\%  & \multicolumn{1}{c|}{15\%}  & \multicolumn{1}{c|}{35\%}  & 55\%  & \multicolumn{1}{c|}{15\%}  & \multicolumn{1}{c|}{35\%}  & 55\%  \\ \hline
RST-S \cite{bai2021dual}                 & \multicolumn{1}{c|}{75.02} & \multicolumn{1}{c|}{76.38} & 76.48 & \multicolumn{1}{c|}{72.90} & \multicolumn{1}{c|}{76.78} & 77.30 & \multicolumn{1}{c|}{78.56} & \multicolumn{1}{c|}{81.18} & 82.19 \\
Group-SL \cite{fang2023depgraph} & \multicolumn{1}{c|}{49.04} & \multicolumn{1}{c|}{77.90} & 78.37 & \multicolumn{1}{c|}{1.43}  & \multicolumn{1}{c|}{4.90}  & 26.24 & \multicolumn{1}{c|}{42.41} & \multicolumn{1}{c|}{67.71} & 79.59 \\
OTOv2 \cite{chen2023otov2}& \multicolumn{1}{c|}{77.04} & \multicolumn{1}{c|}{77.65} & 78.35 & \multicolumn{1}{c|}{76.29} & \multicolumn{1}{c|}{77.35} & 78.39 & \multicolumn{1}{c|}{77.26} & \multicolumn{1}{c|}{80.61} & 80.84 \\
Refill \cite{chen2022coarsening}& \multicolumn{1}{c|}{75.12} & \multicolumn{1}{c|}{77.43} & 78.19 & \multicolumn{1}{c|}{69.57} & \multicolumn{1}{c|}{75.91} & 76.96 & \multicolumn{1}{c|}{75.98} & \multicolumn{1}{c|}{79.25} & 79.56 \\ \hline
Ours                    & \multicolumn{1}{c|}{\textbf{79.64}} & \multicolumn{1}{c|}{\textbf{80.0}}  & \textbf{80.41}& \multicolumn{1}{c|}{\textbf{76.81}} & \multicolumn{1}{c|}{\textbf{78.57}} & \textbf{79.09}& \multicolumn{1}{c|}{\textbf{80.12}} & \multicolumn{1}{c|}{\textbf{81.63}} & \textbf{82.72}\\ \hline
\end{tabular}}
\label{tab:cifar100}
\end{table*}

\begin{table*}[t]
\centering
\caption{Comparison of different pruning methods on Tiny ImageNet. We report the Top-1 accuracy(\%) of dense and pruned networks with different remaining FLOPs.}
\setlength\tabcolsep{10pt}
\resizebox{\linewidth}{!}{
\begin{tabular}{l|ccc|ccc|ccc}
\hline
\multirow{2}{*}{Method} & \multicolumn{3}{c|}{ResNet-50 (Acc: 64.28)}                       & \multicolumn{3}{c|}{MBV3 (Acc: 63.91)}                           & \multicolumn{3}{c}{WRN28-10 (Acc: 61.72)}                          \\ \cline{2-10} 
                        & \multicolumn{1}{c|}{15\%}  & \multicolumn{1}{c|}{35\%}  & 55\%  & \multicolumn{1}{c|}{15\%}  & \multicolumn{1}{c|}{35\%}  & 55\%  & \multicolumn{1}{c|}{15\%}  & \multicolumn{1}{c|}{35\%}  & 55\%  \\ \hline
RST-S \cite{bai2021dual}                 & \multicolumn{1}{c|}{63.03} & \multicolumn{1}{c|}{63.24} & 64.78 & \multicolumn{1}{c|}{55.13} & \multicolumn{1}{c|}{61.26} & 62.76 & \multicolumn{1}{c|}{58.03} & \multicolumn{1}{c|}{61.41} & 62.12 \\
Group-SL \cite{fang2023depgraph}& \multicolumn{1}{c|}{0.95}  & \multicolumn{1}{c|}{19.94} & 55.49 & \multicolumn{1}{c|}{0.56}  & \multicolumn{1}{c|}{2.35}  & 53.43   & \multicolumn{1}{c|}{0.85}  & \multicolumn{1}{c|}{25.74} & 57.64 \\
OTOv2 \cite{chen2023otov2}& \multicolumn{1}{c|}{60.38} & \multicolumn{1}{c|}{63.45} & 65.16 & \multicolumn{1}{c|}{57.61} & \multicolumn{1}{c|}{59.25} & 60.16   & \multicolumn{1}{c|}{57.19} & \multicolumn{1}{c|}{61.23} & 61.70 \\
Refill \cite{chen2022coarsening}& \multicolumn{1}{c|}{61.05} & \multicolumn{1}{c|}{64.14} & 65.02 & \multicolumn{1}{c|}{53.87} & \multicolumn{1}{c|}{61.84} & 62.49 & \multicolumn{1}{c|}{56.64} & \multicolumn{1}{c|}{61.83} & 62.22\\ \hline
Ours                    & \multicolumn{1}{c|}{\textbf{65.93}} & \multicolumn{1}{c|}{\textbf{66.65}} & \textbf{68.27}& \multicolumn{1}{c|}{\textbf{59.74}} & \multicolumn{1}{c|}{\textbf{62.11}} & \textbf{63.64}& \multicolumn{1}{c|}{\textbf{60.53}} & \multicolumn{1}{c|}{\textbf{62.44}} & \textbf{62.81}\\ \hline
\end{tabular}}
\label{tab:timg}
\vspace{-10pt}
\end{table*}

\begin{table*}[t]
\centering
\caption{Verifications of transformers on CIFAR-100. We report the Top-1 accuracy(\%) of dense and pruned networks with different remaining FLOPs.}
\setlength\tabcolsep{10pt}
\resizebox{\linewidth}{!}{
\begin{tabular}{l|c|ccc|ccc}
\hline
                        &                            & \multicolumn{3}{c|}{RST-S~\cite{bai2021dual}}        & \multicolumn{3}{c}{Ours}                                                                                                    \\ \cline{3-8} 
\multirow{-2}{*}{Model} & \multirow{-2}{*}{Dense} & \multicolumn{1}{c|}{15\%}  & \multicolumn{1}{c|}{35\%}  & 55\%  & \multicolumn{1}{c|}{15\%}                         & \multicolumn{1}{c|}{35\%}                         & 55\%                         \\ \hline
ViT                     & 76.49                      & \multicolumn{1}{c|}{70.74} & \multicolumn{1}{c|}{72.05} & 74.65 & \multicolumn{1}{c|}{\textbf{71.28}}                        & \multicolumn{1}{c|}{\textbf{74.49}}                        & \textbf{76.24}\\
Swin                    & 77.16                      & \multicolumn{1}{c|}{70.53}   & \multicolumn{1}{c|}{72.98}   & 75.25& \multicolumn{1}{c|}{\textbf{75.63}} & \multicolumn{1}{c|}{\textbf{76.89}} & \textbf{77.14}\\ \hline
\end{tabular}
}
\label{tab:cifar100-transformer}
\vspace{-10pt}
\end{table*}

\section{Experiment}
In this section, we first validate the effectiveness of the proposed STP on three benchmarks: CIFAR-100~\cite{krizhevsky2009learning}, Tiny ImagneNet~\cite{deng2009imagenet} and ImageNet~\cite{deng2009imagenet}. For CIFAR100 and Tiny ImageNet, three typical CNNs, including ResNet-50~\cite{he2016deep}, MobileNetV3 (MBV3)~\cite{howard2019searching} and WRN28-10~\cite{zagoruyko2016wide}, are evaluated under different pruning ratios, \ie, 15\%, 35\% and 55\%. For ImageNet, we choose ResNet-50 as the backbone and compare STP with various structured pruning methods on Top-1 accuracy and FLOPs. Following the benchmarking, ablation studies are conducted to verify the indispensability of each component. Finally, we analyze the KD guided exploration process and the parameter magnitude empirically to reveal the internal mechanism of STP. All ablation and investigation experiments are conducted on CIFAR-100 with ResNet-50. We adopt the standard training settings without bells and whistles. Details can be referred to in Appendix.
\subsection{Image Classifcation}
\noindent \textbf{Results on CIFAR-100 and Tiny ImageNet.} To verify the effectiveness of STP and show its robustness to different networks. We conduct experiments on CIFAR-100 and Tiny ImageNet using ResNet-50, MBV3, and WRN28-10 as backbones. For each benchmark-network pair, three different FLOPs ratios are considered, \ie, 15\%, 35\%, and 55\%. The leading methods for comparison include a slightly modified structured RST~\cite{bai2021dual} (named RST-S\footnote{The original RST does not support a structured ticket, we keep its regular training loss unaltered while constraining the randomly picked subnetwork to be structured.}), Group-SL~\cite{fang2023depgraph}, OTOv2~\cite{chen2023otov2}, and Refill~\cite{chen2022coarsening}. For fair comparisons, we utilize consistent basic training settings for STP and the others. The results are shown in Table~\ref{tab:cifar100} and Table~\ref{tab:timg}. Compared with the others, STP consistently achieves state-of-the-art performance, especially in extremely low FLOPs. For example, when reducing the FLOPs of ResNet-50 to 15\%, Group-SL suffers from a severe accuracy drop from 77.49\% (35\%) to 49.04\% (15\%). In contrast, STP still maintains high accuracy, and the accuracy gains compared with the second-best method are up to 2.6\% on CIFAR-100 and 2.9\% on Tiny ImageNet, respectively.
\par To further demonstrate the generality of STP, we apply STP on two typical Transformers, i.e., ViT~\cite{vaswani2017attention} and Swin Transformer~\cite{liu2021Swin}. Similar to CNNs, we conduct experiments on CIFAR-100 w.r.t. three FLOPs targets, including 15\%, 35\%, and 55\%. The results are shown in Table~\ref{tab:cifar100-transformer}. It can be observed that STP is superior for both Transformers under all target FLOPs compared with RST-S. Moreover, under 55\% FLOPs, the STP pruned Transformers can almost realize lossless compression, suffering merely 0.25 and 0.02 performance drop for ViT and Swin Transformer, respectively. Note that the proposed STP is not tailored for Transformers, however, it can still achieve competitive results under 55\% FLOPs, implying its immense potential in pruning Transformers. 

\begin{table}[ht]
\centering
\vspace{-10pt}
\caption{Results of ResNet-50 on Imagenet. We report the Top-1 accuracy(\%) of dense and pruned networks with different remaining FLOPs.}
\resizebox{\linewidth}{!}{
\begin{tabular}{lcccc}
\hline
Method        & Unpruned top-1 (\%) & Pruned top-1 (\%) & Top-1 drop (\%) & FLOPs (\%)     \\ \hline
OTOv2~\cite{chen2023otov2}         & 76.10               & 70.10             & 6.00            & \textbf{14.50} \\
Refill~\cite{chen2022coarsening}        & 75.84               & 66.83             & 9.01            & 20.00          \\
\textbf{Ours} & 76.15               & \textbf{72.43}    & \textbf{3.72}   & 15.26          \\ \hdashline
MetaPruning~\cite{liu2019metapruning}   & 76.60               & 73.40             & 3.20            & 24.39          \\
Slimmable~\cite{yu2018slimmable}     & 76.10               & 72.10             & 4.00            & 26.63          \\
ThiNet~\cite{luo2017thinet}        & 72.88               & 68.42             & 4.46            & 28.50          \\
OTOv2~\cite{chen2023otov2}         & 76.10               & 74.30             & 1.80            & 28.70          \\
GReg-1~\cite{wang2021neural}        & 76.13               & 73.75             & 2.38            & 32.68          \\
GReg-2~\cite{wang2021neural}        & 76.13               & 73.90             & 2.23            & 32.68          \\
CAIE~\cite{wu2020constraint}          & 76.13               & 72.39             & 3.74            & 32.90          \\
\textbf{Ours} & 76.15               & \textbf{75.39}    & \textbf{0.76}   & \textbf{24.21} \\ \hdashline
CHIP~\cite{sui2021chip}          & 76.15               & 75.26             & 0.89            & 37.20          \\
OTOv2~\cite{chen2023otov2}
& 76.10               & 75.20             & 0.90            & 37.30          \\
GReg-1~\cite{wang2021neural}        & 76.13               & 74.85             & 1.28            & 39.06          \\
GReg-2~\cite{wang2021neural}        & 76.13               & 74.93             & 1.20            & 39.06          \\
Refill~\cite{chen2022coarsening}        & 75.84               & 72.25             & 3.59            & 40.00          \\
ThiNet~\cite{luo2017thinet}        & 72.88               & 71.01             & 1.87            & 44.17          \\
GBN~\cite{you2019gate}           & 75.85               & 75.18             & 0.67            & 44.94          \\
FPGM~\cite{he2019filter}          & 76.15               & 74.83             & 1.32            & 46.50          \\
LeGR~\cite{chin2020towards}          & 76.10               & 75.30             & 0.80            & 47.00          \\
AutoSlim~\cite{yu2019autoslim}     & 76.10               & 75.60             & 0.50            & 48.43          \\
MetaPruning~\cite{liu2019metapruning}   & 76.60               & 75.40             & 1.20            & 48.78          \\
\textbf{Ours} & 76.15               & \textbf{75.68}    & \textbf{0.47}   & \textbf{35.00} \\ \hdashline
CAIE~\cite{wu2020constraint}          & 76.13               & 75.62             & 0.51            & 54.77          \\
CHIP~\cite{sui2021chip}          & 76.15               & 76.30             & -0.15           & 55.20          \\
Slimmable~\cite{yu2018slimmable}     & 76.10               & 74.90             & 1.20            & 55.69          \\
TAS~\cite{dong2019network}           & 77.46               & 76.20             & 1.26            & 56.50          \\
SSS~\cite{huang2018data}           & 76.12               & 71.82             & 4.30            & 56.96          \\
FPGM~\cite{he2019filter}          & 76.15               & 75.59             & 0.56            & 57.80          \\
LeGR~\cite{chin2020towards}          & 76.10               & 75.70             & 0.40            & 58.00          \\
GBN~\cite{you2019gate}           & 75.88               & 76.19             & -0.31           & 59.46          \\
Refill~\cite{chen2022coarsening}        & 75.84               & 74.46             & 1.38            & 60.00          \\
ThiNet~\cite{luo2017thinet}       & 72.88               & 72.04             & 0.84            & 63.21          \\
GReg-1~\cite{wang2021neural}        & 76.13               & 76.27             & -0.14           & 67.11          \\
MetaPruning~\cite{liu2019metapruning}   & 76.60               & 76.20             & 0.40            & 73.17          \\
\textbf{Ours} & 76.15               & \textbf{77.13}    & \textbf{-0.98}  & \textbf{53.49} \\ \hdashline
SSS~\cite{huang2018data}           & 76.12               & 75.44             & 0.68            & 84.94          \\
\textbf{Ours} & 76.15               & \textbf{77.48}    & \textbf{-1.33}  & \textbf{75.00} \\ \hline
\end{tabular}
}
\vspace{-30pt}
\label{tab:imagenet}
\end{table}

\noindent \textbf{Results on ImageNet.} We also validate STP on the widely used ImageNet-1K benchmark, reporting the performance of STP-pruned ResNet-50 under various FLOPs ranging from 15\% to 75\% and comparing them with the ones pruned by other methods. As shown in Table.~\ref{tab:imagenet}, under nearly the same FLOPs, the network pruned by STP consistently achieves higher accuracy than the ones pruned by other methods, demonstrating the superiority of STP. Considering extremely low FLOPs interval (from 10\% to 20\%), few methods (STP, OTOv2, and Refill) attempt to prune ResNet-50 to this interval. Among these methods, our STP achieves the highest 72.43\% top-1 accuracy, overpassing OTOv2 by 2.33\% while remaining almost the same FLOPs (around 15\%).

\subsection{Ablation Studies}
We eliminate the proposed components of STP, \ie, ST relative sparsification, subnet mutating expansion, KD guided architecture exploration and multi-dimension sampling, to validate the single and joint effect of them. The results are shown in Table~\ref{tab:ab}. Firstly, we conduct experiments with the separate removal of each component except the ST relative sparsification and observe different degrees of performance degradation, which demonstrates their indispensability. Then, we remove these components step-by-step and observe consistent performance degradation, which demonstrates their joint effectiveness. Note that, to remove the ST relative sparsification, we replace it with the $L_2$ regularization, which causes significant performance damage (-2.7\% Top-1 accuracy), demonstrating the superiority of the ST relative sparsification.

\begin{table}[]
\small
\centering
\vspace{-10pt}
\caption{Influence of different components in stimulative training guided pruning.}
\resizebox{\linewidth}{!}{
\begin{tabular}{cccc|c}
\hline
\begin{tabular}[c]{@{}c@{}}ST relative \\ sparsification\end{tabular} & \begin{tabular}[c]{@{}c@{}}Subnet mutating \\ expansion\end{tabular} & \begin{tabular}[c]{@{}c@{}}KD guided \\ architecture exploration\end{tabular} & \begin{tabular}[c]{@{}c@{}}Multi-dimension \\ sampling\end{tabular} & Top-1 Acc(\%) \\ \hline
\checkmark                                             & \checkmark                                            & \checkmark                                                                         & \checkmark                                           & \textbf{79.64}         \\
\checkmark                                             & ×                                                                    & \checkmark                                                                         & \checkmark                                           & 78.12         \\
\checkmark                                             & \checkmark                                            & ×                                                                                                 & \checkmark                                           & 78.47         \\
\checkmark                                             & \checkmark                                            & \checkmark                                                                         & ×                                                                   & 78.23         \\
\checkmark                                             & \checkmark                                            & ×                                                                                                 & ×                                                                   & 77.97         \\
\checkmark                                             & ×                                                                    & ×                                                                                                 & ×                                                                   & 77.46         \\
×                                                                     & ×                                                                    & ×                                                                                                 & ×                                                                   & 74.76         \\ \hline
\end{tabular}

}
\label{tab:ab}
\vspace{-30pt}
\end{table}

\begin{table*}[t]
\small
\centering
\caption{The results of downstream tasks, i.e., object detection and instance segmentation, on COCO2017. The pretrained pruned ResNet-50 with different remaining FLOPs are utilized as the backbone in Mask R-CNN and fine-tuned on different tasks.}
\renewcommand{\arraystretch}{1.2}
\resizebox{\linewidth}{!}{
\begin{tabular}{l|c|ccc|ccc}
\hline
Method                & ImageNet Acc (\%) & Det mAP & Det mAP@0.75  & Det mAP@0.5 & Seg mAP  & Seg mAP@0.75  & Seg mAP@0.5   \\ \hline
Baseline (100\% FLOPs) & 76.15   & 37.8         & 40.8          & 58.6    & 34.5      & 36.7          & 55.4          \\ \hline
GReg-2~\cite{wang2021neural} (40\% FLOPs) & 74.93 & 36.6 & 39.6 & 57.4 & 33.9 & 36.2 & 54.4 \\     
GReg-1~\cite{wang2021neural} (67\% FLOPs) & 76.27 & 37.6 & 40.5 & 58.7 & 34.6 & 36.4 & 55.3 \\ \hline
Ours (15\% FLOPs)       & 72.43     & 35.0       & 37.7          & 55.6    & 32.5      & 34.6          & 52.8          \\
Ours (35\% FLOPs)       & 75.68     & 37.4       & 40.7          & 58.1    & 34.5     & 36.6          & 55.2          \\
Ours (55\% FLOPs)       & \textbf{77.13}  & \textbf{38.1} & \textbf{41.1} & \textbf{59.0} & \textbf{35.1} & \textbf{37.6} & \textbf{55.8} \\ \hline
\end{tabular}}
\vspace{-15pt}
\label{tab:downstream_tasks}
\end{table*}

\subsection{Downstream Tasks}
To validate the practical capacity of STP, we conduct experiments on downstream tasks, including object detection and instance
segmentation. Specifically, we apply STP on ResNet-50 in ImageNet to generate pretrained pruned networks with three remaining FLOPs, i.e., 15\%, 35\%, 55\%. These pretrained pruned networks are utilized as the backbone in Mask R-CNN~\cite{he2017mask}, which is a multi-task framework. Then, the whole Mask R-CNN is fine-tuned and trained on COCO2017~\cite{lin2014microsoft} under the supervision of object detection and instance segmentation. For the fine-tuning settings, we follow the standard 1x training schedule~\cite{chen2019mmdetection} and adopt AdamW~\cite{loshchilov2019decoupled} to train the Mask R-CNN for 12 epochs with weight decay 0.1. The batch size is 16 and the initial learning rate is 0.0001. The learning rate decays 10\% at the 8-th and the 12-th epoch. We report the mean average precision (mAP) under different intersection over union (IoU) of the bounding boxe and mask as the evaluation metrics of object detection and instance segmentation, respectively.
\par The results are shown in Table~\ref{tab:downstream_tasks}. It can be observed that STP consistently surpasses GReg~\cite{wang2021neural} on both detection and segmentation. Under relatively low FLOPs (35\%), STP causes marginal performance reduction (-0.1\% on Det mAP@0.75 and Seg mAP@0.75) on downstream tasks. In the case of extremely low FLOPs(15\%), STP can maintain the basic performance and does not cause unbearable performance degradation. It is worth noting that STP can improve performance (+0.3\% on Det mAP@0.75 and +0.9\% on Seg mAP@0.75) with 1.8x speed up (55\% remaining FLOPs). We consider the reason is that the ST guided training framework can drive the pruned network to extract more representative features for downstream tasks. The above observations demonstrate that STP can be simply and practically deployed in different tasks and obtain competitive performance with various target FLOPs, which is friendly for resource-limited edge devices. Besides, the performance improvement suggests the potential of STP to improve pareto frontier between FLOPs and performance.
\subsection{Evaluation of Architecture Exploration}
To verify that our architecture exploration strategy can gradually explore the subnets with outstanding performance, we record the pool in different training epochs. We evaluate the effectiveness from two aspects: 1) performance: the average performance of architectures in the pool 2) clustering: the average sum of squared error (SSE) distance between the clustering center and architectures in the pool, where the higher SSE means lower clustering. Notely, for estimating the clustering of architectures, we convert the normalized sparsity ratio of each layer and depth into a vector and regard it as a proxy of architectures. Due to the unbearable computational overhead of fully training and measuring each subnet in the initialized pool(containing more than 1000 subnets), we conduct another experiment that utilizes the original ST while uniformly sampling the whole pool. After ST, we evaluate all subnets in the initialized pool on the test dataset as the proxy (or a look table) of the architectures' performance. 
\par
\begin{figure}[t]
\centering
\begin{minipage}[t]{0.49\linewidth}
\centering
\includegraphics[width=\linewidth]{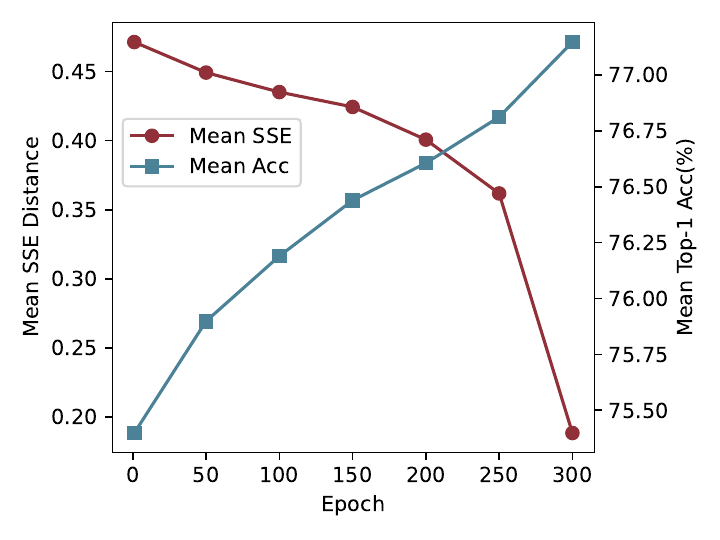}
\caption{Evaluation of KD guided architecture exploration. The mean sum of suqred error (SSE) distance measures the divergence of architectures pool, and lower SSE means higher clustering. During the training, such exploration progressively concentrates on high-performance architectures.}
\label{fig:pool_mean_and_dist}
\end{minipage}
\hfill
\begin{minipage}[t]{0.49\linewidth}
\centering
\includegraphics[width=\linewidth]{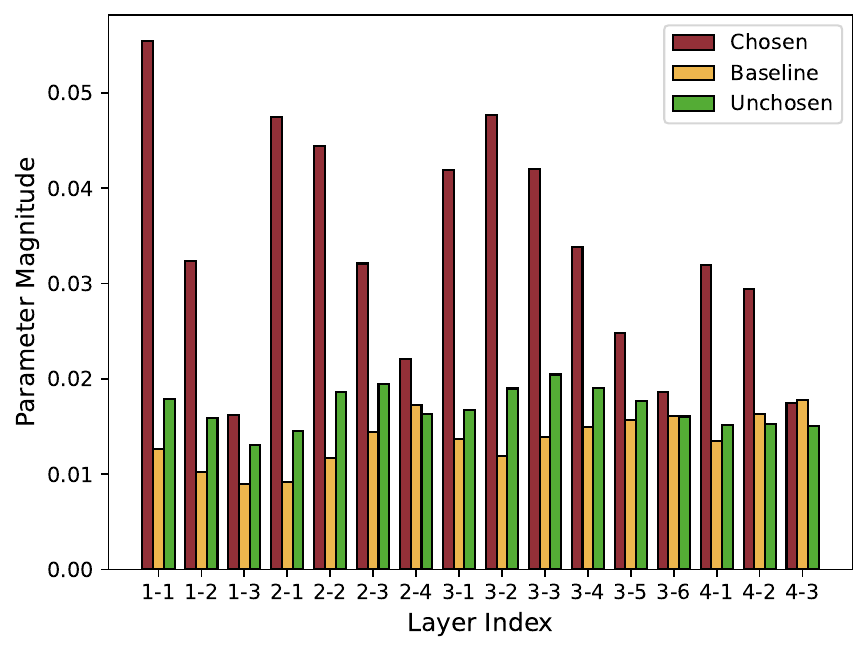}
\caption{The parameter magnitude of the main network trained by STP. The horizontal axis is the layer index like Fig.~\ref{fig:weight_distribution}. It empirically verifies that STP enhances the chosen (kept) parameters with a relative sparsity effect while maintaining the magnitude of unchosen (dropped) ones.}
\label{fig:width_distribution}
\end{minipage}
\vspace{-10pt}
\end{figure}

 The results are shown in Fig.~\ref{fig:pool_mean_and_dist}. At the early stage of STP, the randomly initialized pool contains various architectures with low average performance and markedly different structures. As training progresses, the average performance is gradually raised, and at the same time, the similarity of structures also increases. It demonstrates that under the constraint of FLOPs, our strategy can progressively drive the architecture exploration to converge to an optimal region and finally obtain a high-performance architecture for sparsification.

\subsection{Parameter Magnitude Analysis}
To verify the relative sparsification effect of STP, we record the average magnitude of the chosen (kept), unchosen (dropped), and baseline (standard training without sparsification) parameters in different convolutional layers of ResNet-50. As illustrated in Fig.~\ref{fig:width_distribution}, the magnitude of chosen parameters is remarkably and consistently higher than the unchosen part in the front layers, while the unchosen part has a similar magnitude as the baseline. It demonstrates that STP achieves the evident relative sparsity effect like the original ST without suppressing the dropped part. Due to the subnet mutating expansion, the unchosen part has small parability of being enhanced, thus, the magnitude of unchosen part is marginally higher than the baseline. There is another phenomenon that in the later layer of stages, the relative sparsification effect becomes less pronounced. That is because during training, for low FLOPs targets, e.g., 15\%, the later layers are dropped with a greater probability and sparsified less. Combining with the empirical results in Table~\ref{tab:cifar100}, we conclude that STP can obtain a sparsity parameter distribution for pruning with less performance degradation.
\section{Conclusion}
Suppressed sparsification paradigm has been widely utilized in sparsification-based pruning, but suffers from capacity damage before pruning. In this manuscript, we reveal the relative sparsity effect in ST. Based on it, we propose an enhanced sparsification paradigm for structured pruning. Besides, we propose the multi-dimension sampling space, subnet mutating extension, and architecture pool exploration strategy, which formulate a one-stage pruning framework, named stimulative training guided pruning (STP), which can obtain a compact network with extremely low FLOPs and high performance. Comprehensive experiments on mainstream datasets and networks verify the effectiveness of STP.
%
%
\bibliographystyle{splncs04}
\bibliography{main}
\vspace{150pt}
\section*{Appendix A: Implementation Details}
In this section, we introduce the detailed training settings of experiments in the main manuscript. All experiments are implemented using Pytorch~\cite{pytorch2019pytorch}.
\subsection*{A1. CIFAR-100 implementation details}
The CIFAR-100~\cite{krizhevsky2009learning} is a typical classification dataset with 100 categories, consisting of 50,000 training images and 10,000 testing images. For ResNet-50~\cite{he2016deep} and MBV3~\cite{howard2019searching}, we adopt the training settings of~\cite{ye2022stimulative}. Specifically, the epoch number and batch size are 500 and 64, respectively. The SGD is chosen as the optimizer with a 0.05 initial learning rate and a 0.0003 weight decay. We use the cosine decay schedule to adjust the learning rate over the training process. For WRN28-10~\cite{zagoruyko2016wide}, we adopt the training settings of~\cite{zagoruyko2016wide}. Specifically, the epoch number and batch size are 200 and 128, respectively. The SGD is chosen as the optimizer with a 0.1 initial learning rate and a 0.0005 weight decay. The learning rate scheduler is also the cosine decay schedule. For ViT~\cite{vaswani2017attention} and Swin Transformer~\cite{liu2021Swin}, we use an image size of 32x32 and a patch size of 4. The epoch number and batch size are 200 and 128, respectively. The optimizer is AdamW~\cite{loshchilov2019decoupled} with an initial learning rate of 0.001/0.003 for Swin/ViT and a 0.05 weight decay. The learning rate is warmed up for 10 epochs. The data augmentations are the same as the ones in \cite{lee2021vision}. Different from CNNs, where we regard the channel numbers of convolutional and linear layers as the width dimension, to prune the width of Transformers, we take the head numbers of attention layers and the channel numbers of linear layers into account.  
\subsection*{A2. Tiny ImageNet implementation details}
The Tiny Imagenet dataset is inherited from the ImageNet dataset~\cite{deng2009imagenet}, containing 200 categories, 100,000 training images, and 10,000 testing images.  For ResNet-50~\cite{he2016deep} and MBV3~\cite{howard2019searching}, the epoch number and batch size are 500 and 64, respectively. The optimizer is SGD with a 0.1 initial learning rate and a 0.0003 weight decay. We utilize a step-wise learning rate scheduler, downscaling the learning rate to 0.1 and 0.01 of the original one at the 250-th and 375-th epoch, respectively.
For WRN28-10~\cite{zagoruyko2016wide}, we adopt the training settings in~\cite{shen2022self}. The epoch number and batch size are 200 and 128, respectively. The optimizer is SGD with a 0.2 initial learning rate and a 0.0001 weight decay. We utilize a step-wise learning rate scheduler, downscaling the learning rate to 0.1 and 0.01 of the original one at the 100-th and 150-th epoch, respectively.
\subsection*{A3. ImageNet implementation details}
The ImageNet dataset~\cite{deng2009imagenet} is a widely used classification benchmark, containing 1000 categories, 1.2 million training images, and 50000 testing images. For the evaluated ResNet-50~\cite{he2016deep}, the epoch number and batch size are 200 and 512, respectively. We utilize SGD as the optimizer. The learning rate is initialized as 0.2 and is controlled by a cosine decay schedule. The weight decay is 0.0001. Besides, we apply the commonly used data augmentations according to~\cite{huang2017densely,szegedy2015going}.
\subsection*{A5. Pruning framework details}
To build a relatively more general and user-friendly pruning framework, we resort to the symbolic tracing of Pytorch, called FX tracing~\cite{pytorch2022fx}. Given any FX-compatible model, we first extract the FX graph from it. Based on the graph, we try to extract dependencies groups, which are similar to the concepts in~\cite{fang2023depgraph,chen2023otov2}. A dependency group represents a series of layers whose outputs are expected to be integrated, such as being added or concatenated in the dimension to prune. When pruned, all layers in the same group are expected to have the same number of remaining channels. 

Given the inputs to the model, we run an interpreter~\cite{pytorch2022fx} based on the extracted FX graph instead of using the traditional model forward. The interpreter propagates the inputs from the root node of the graph to the last node, executing all encountered nodes according to their recorded operations, promising the same result as the one produced by the traditional model forward while providing the freedom to intercept any intermediate operation for injecting necessary functionality of pruning. Benefiting from the interpreter, we can dynamically reduce the layer number or shrink channels based on the sampled architecture without altering the model structure or parameters. To realize the dynamic interpreter, we dispatch all commonly used operators, such as trivial binary operators (add, mul, div, etc.), Pytorch tensor operators, and ``torch.nn" forward functions, in the FX graph to a pre-registered dynamic handler, which can be easily extended to support customized operations. 

Besides dynamic forward, we closely integrate FLOPs/parameters estimation with the above pruning framework via running an interpreter with a proxy tensor. A proxy tensor does not require dense calculation, such as convolution or matrix multiplication, resulting in significant acceleration of the FLOPs/parameters estimation for dynamic models, which is the key to estimating the FLOPs of all architectures within the pool in a bearable time. During an interpreter run, a proxy tensor only contains the shape information of a tensor, and each encountered operation is required to infer the output shape of the tensor in addition to the FLOPs and parameters introduced by this operation. Similar to the dispatch of dynamic forward, we have registered the handlers of FLOPs/parameters estimation for all commonly used operators in the FX graph.

\section*{Appendix B: Pseudo Code of STP}
To understand the detailed process of STP, the pseudo code of STP is shown in Alg.~\ref{algorithm}. Some expressions in Alg.~\ref{algorithm} are different from the manuscript and they are explained as follows to avoid confusion.
\par In Alg.~\ref{algorithm}, we denote ``$\leftarrow$" as the assignment operator and ``$=$" as the equal comparison operator. Different from the sparsity ratio $S_g$, we utilize the target remaining FLOPs ratio $r$, which can achieve more precise control over the inference speed of the pruned network. $N_{shr} \leftarrow \left \lfloor \frac{k(N_p - 1)}{T_{shr}} \right \rfloor $ is the removing number of architectures in each shrinking process, meaning that after shrinking $\frac{T_{shr}}{k}$ rounds, i.e., the $T$ in Section~3.3 and Fig.~5 in the manuscript, the number of architectures in the pool $\mathcal{P}$ will reduce from $N_p$ to 1. Notely, the rounding down of $\frac{k(N_p - 1)}{T_{shr}}$ can cause that there are several architectures in $\mathcal{P}$; thus, we force to shrink the number of architecture in $\mathcal{P}$ into 1. The $\mathcal{N}_{s}$ and $\mathcal{N}_{sup}$ are the ``Sub Network" and ``Mut Network" in Fig.~5 respectively, while the $\mathcal{L}_{STC}$ and $\mathcal{L}_{SME}$ are the ``$KD_{sub}$" and ``$KD_{mut}$" in Fig.~5 respectively. $\beta_1$ and $\beta_2$ are the loss coefficients of $\mathcal{L}_{STS}$ and $\mathcal{L}_{SME}$ respectively.    

\begin{algorithm*}
\caption{Stimulative training guided pruning (STP)}
\label{algorithm}
\KwIn{A main network $\mathcal{N}_m$ with randomly initialized parameters $\theta_m$; target remaining FLOPs ratio $r$; total training steps $T_{total}$; initial pool size $N_p$, the end steps of shrinking pool $T_{shr}$, pool refine interval $k$ and shrinking number $N_{shr} \leftarrow \left \lfloor \frac{k(N_p - 1)}{T_{shr}} \right \rfloor $ of each $k$ steps; Input $x$ and ground truth $y$ of each minibatch;}
\KwOut{A pruned network $\mathcal{N}_s^*$ satisfying the target FLOPs ratio $r$ with parameters $\theta_s^*$;}
$\rhd$ Construct the architecture pool $\mathcal{P}$ by randomly sampling $N_p$ subnets that satisfy target FLOPs ratio $r$; Initialize the training step $t \leftarrow 1$;

\While{$t \le T_{total}$}
{
$\rhd$ forward $N_m$ to obtain the output $Z_m(x;\theta_m)$ and compute the cross entropy loss with the ground truth $\mathcal{L}_{CE}$;

$\rhd$ Sample a target subnet $\mathcal{N}_s$ from architecture pool $\mathcal{P}$, forward it and compute KL- loss $\mathcal{L}_{STS}$ with $Z_m(x;\theta_m)$ as Formulation~4; record $\mathcal{L}_{STS}$ and update the corresponding score in $\mathcal{P}$ as Formulation~8;

$\rhd$ Mutate and expand the target subnet $\mathcal{N}_s$ to obtain $\mathcal{N}_{sup}$, forward $\mathcal{N}_{sup}$ and compute KL- loss $\mathcal{L}_{SME}$ with $Z_m(x;\theta_m)$ as Formulation~7;

$\rhd$ Backward the total loss $\mathcal{L}_{total} \leftarrow \mathcal{L}_{CE} + \beta_1 \mathcal{L}_{STS} + \beta_2 \mathcal{L}_{SME}$ and update $\theta_m$ with by descending $\nabla_{\theta_m}\mathcal{L}_{total}$

\If{$t \; mod \; k = 0$ and $t \le T_{shr}$}{
$\rhd$ remove $N_{shr}$ architectures with Top-$N_{shr}$ score in the pool $\mathcal{P}$;
}
$\rhd$ $t \leftarrow t + 1$;
}
$\rhd$ There is only one architecture in the pool $\mathcal{P}$, extract it from $\mathcal{N}_m$ and regard it as the pruned network $\mathcal{N}_s^*$ with parameters $\theta_s^*$.
\end{algorithm*}

\begin{figure*}[t]
  \centering
  \includegraphics[width=\linewidth]{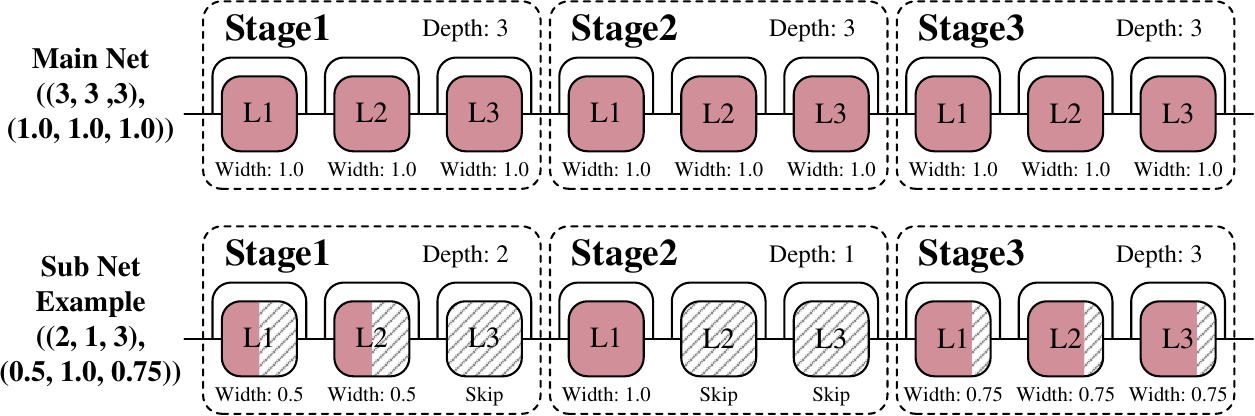}
  \caption{The visualization of sampled architectures for an exemplary network. The exemplary network is composed of three stages. Each stage contains three layers that can be pruned, such as the convolutional layers.}
  \label{fig:proxy}
\end{figure*}

\begin{table*}[]
\small
\centering
\setlength{\tabcolsep}{5.7pt}
\renewcommand{\arraystretch}{1.2}
\caption{The remaining subnet architectures of ResNet-50 in the pool when the pool size is reduced to 10 under different FLOPs targets during the training on CIFAR-100. The Top-1 accuracy (``Acc (\%)") of each architecture is measured after training. The architecture in bold is the final one remaining in the pool.}
\resizebox{\linewidth}{!}{
\begin{tabular}{cc|cc|cc}
\hline
\multicolumn{2}{c|}{15\% FLOPs Pool}                                      & \multicolumn{2}{c|}{35\% FLOPs Pool}                                      & \multicolumn{2}{c}{55\% FLOPs Pool}                                       \\ \hline
\multicolumn{1}{c|}{Architecture}                         & Acc (\%) & \multicolumn{1}{c|}{Architecture}                         & Acc (\%) & \multicolumn{1}{c|}{Architecture}                         & Acc (\%) \\ \hline
\multicolumn{1}{c|}{((1, 2, 5, 2), (0.5, 0.3, 0.3, 0.7))} & 79.51    & \multicolumn{1}{c|}{((1, 2, 4, 3), (0.3, 0.7, 0.5, 1.0))} & 79.79    & \multicolumn{1}{c|}{((3, 2, 5, 3), (0.3, 0.7, 0.9, 1.0))} & 80.16    \\
\multicolumn{1}{c|}{((1, 3, 6, 2), (0.3, 0.3, 0.3, 0.7))} & 79.63    & \multicolumn{1}{c|}{((2, 3, 3, 3), (0.3, 0.3, 0.7, 1.0))} & 79.86    & \multicolumn{1}{c|}{((3, 2, 5, 3), (0.5, 0.5, 0.9, 1.0))} & 80.13    \\
\multicolumn{1}{c|}{((1, 3, 4, 2), (0.5, 0.3, 0.3, 0.7))} & 79.63    & \multicolumn{1}{c|}{((3, 4, 5, 3), (0.3, 0.3, 0.5, 1.0))} & 79.97    & \multicolumn{1}{c|}{((3, 2, 6, 3), (0.3, 0.3, 0.9, 1.0))} & 80.21    \\
\multicolumn{1}{c|}{\textbf{((2, 3, 4, 2), (0.3, 0.3, 0.3, 0.7))}} & \textbf{79.64}& \multicolumn{1}{c|}{((3, 2, 6, 3), (0.3, 0.3, 0.5, 1.0))} & 79.74    & \multicolumn{1}{c|}{((2, 2, 5, 3), (0.5, 0.5, 0.9, 1.0))} & 80.14    \\
\multicolumn{1}{c|}{((1, 3, 5, 2), (0.3, 0.3, 0.3, 0.7))} & 79.48    & \multicolumn{1}{c|}{((1, 2, 3, 3), (0.5, 0.3, 0.7, 1.0))} & 79.69    & \multicolumn{1}{c|}{\textbf{((2, 4, 5, 3), (0.5, 0.5, 0.9, 0.9))}} & \textbf{80.41}\\
\multicolumn{1}{c|}{((1, 4, 6, 2), (0.3, 0.3, 0.3, 0.7))} & 79.54    & \multicolumn{1}{c|}{((3, 2, 4, 3), (0.5, 0.3, 0.5, 1.0))} & 79.56    & \multicolumn{1}{c|}{((1, 4, 6, 3), (0.3, 0.3, 0.9, 1.0))} & 80.07    \\
\multicolumn{1}{c|}{((1, 3, 5, 2), (0.5, 0.3, 0.3, 0.7))} & 79.39    & \multicolumn{1}{c|}{((1, 2, 4, 3), (0.9, 0.3, 0.5, 1.0))} & 79.59    & \multicolumn{1}{c|}{((2, 2, 6, 3), (0.3, 0.3, 0.9, 1.0))} & 80.08    \\
\multicolumn{1}{c|}{((1, 2, 3, 3), (0.3, 0.3, 0.3, 0.7))} & 79.16    & \multicolumn{1}{c|}{\textbf{((1, 4, 3, 3), (0.5, 0.3, 0.7, 1.0))}} & \textbf{80.0}& \multicolumn{1}{c|}{((2, 3, 5, 3), (0.5, 0.5, 0.9, 1.0))} & 80.07    \\
\multicolumn{1}{c|}{((1, 1, 4, 3), (0.3, 0.3, 0.3, 0.7))} & 79.31    & \multicolumn{1}{c|}{((3, 2, 5, 3), (0.3, 0.5, 0.5, 1.0))} & 79.82    & \multicolumn{1}{c|}{((1, 4, 5, 3), (0.5, 0.3, 1.0, 0.9))} & 80.05    \\
\multicolumn{1}{c|}{((2, 1, 3, 3), (0.3, 0.3, 0.3, 0.7))} & 79.05    & \multicolumn{1}{c|}{((1, 4, 3, 3), (0.3, 0.3, 0.7, 1.0))} & 79.78    & \multicolumn{1}{c|}{((3, 2, 5, 3), (0.5, 0.7, 0.9, 0.9))} & 80.01    \\ \hline
\end{tabular}
}
\label{tab:proxy}
\end{table*}

\begin{table*}[]
\small
\centering
\renewcommand{\arraystretch}{1.2}
\caption{The final remaining architecture in repeated experiments (``Exp") using different seeds for ResNet-50 on CIFAR100 with a 15\% FLOPs target. The Top-1 accuracy (``Acc (\%)"), FLOPs, and parameters (``Params") of the architecture are reported. For FLOPs and parameters, we use their relative fraction w.r.t. the ones of the main network as the metrics.}
\begin{tabular}{l|c|c|c|c}
\hline
Exp & Architecture                         & Acc (\%) & FLOPs (\%) & Params (\%) \\ \hline
\#1 & ((2, 3, 5, 2), (0.3, 0.3, 0.3, 0.7)) & 79.47    & 14.88      & 22.28       \\
\#2 & ((1, 3, 6, 2), (0.3, 0.3, 0.3, 0.7)) & 79.49& 14.69      & 22.36       \\
\#3 & ((1, 2, 5, 2), (0.5, 0.3, 0.3, 0.7)) & 79.55    & 15.22      & 22.33       \\
\#4 & ((2, 3, 4, 2), (0.3, 0.3, 0.3, 0.7)) & 79.64    & 14.89      & 21.94       \\
\#5 & ((2, 2, 6, 2), (0.3, 0.3, 0.3, 0.7)) & 79.57    & 15.23      & 22.93\\ \hline
\end{tabular}
\label{tab:multi_exp_proxy}
\end{table*}

\section*{Appendix C: Exemplary Sampled Subnets}
Assuming an exemplary network with three stages, each consisting of three prunable layers, such as convolutional layers or linear layers, we encode the sampled architecture as a nested tuple as shown in Fig.~\ref{fig:proxy}. The first inner tuple denotes the remaining layers of each stage (the depth dimension), and each element in the second inner tuple denotes the layer-wise proportion of the remaining output channels to the total output channels in the corresponding stage (the width dimension). 

Based on the nested tuple denotation, snapshots of the architecture pool are shown in Table~\ref{tab:proxy}. Given different FLOPs targets, i.e., 15\%, 35\%, and 55\%, we conduct experiments on CIFAR-100 using ResNet-50 as the backbone. The ResNet-50 has four stages, containing 3, 4, 6, and 3 blocks, respectively. A block is composed of three convolutional layers. Slightly different from the former exemplary network, we pack the three convolutional layers and view them as a ``layer" that will be skipped or retained together based on a given depth and will be pruned with the same width. The minimum and maximum widths are set to 0.3 and 1.0, respectively. The width granularity is 0.2, resulting in 5 different choices (0.3, 0.5, 0.7, 0.9, 1.0). It can be observed from Table~\ref{tab:proxy} that 1) when the pool is about to converge (the pool size is reduced to 10), the remaining architectures are similar to each other, especially on the width dimension. For example, except for the first stage, where widths are either 0.3 or 0.5, the widths of the second, the third, and the final stage are consistently 0.3, 0.3, and 0.7, respectively; 2) From the aspect of depth, layers in the last two stages tend to be retained. These observations might help infer the relative importance of different stages; for example, the final stage seems to be the most important one because of relatively more remaining width and depth.

\section*{Appendix D: Robustness of Architecture Pool}
To explore the robustness of the architecture pool, we choose 15\% as the FLOPs target and conduct five experiments with different random seeds on CIFAR-100 using ResNet-50 as the backbone. The results are shown in Table~\ref{tab:multi_exp_proxy}. The final architectures in different experiments are highly similar in accordance with the observations in Appendix C. Besides, the Top-1 accuracy, FLOPs, and parameters of the final architecture are relatively consistent with negligible deviations (within 0.2\% for Top-1 accuracy deviations, within 1\% for FLOPs and parameters deviations), verifying the robustness of the architecture pool, i.e., the converged architecture is relatively stable and its performance is guaranteed.

\section*{Appendix E: Theoretical Explanation and Analysis of Relative Sparsity Effect in ST}
Due to the highly complex topology and non-linearity, explaining the relative sparsity effect in a deep neural network is difficult. Rather than giving completed and rigorous proof, we provide an explanation of a degenerate case for an intuitive understanding. We analyze the ST process, which transfers the capacity from the whole parameters to the chosen ones.
\par For simplicity, we consider a single-layer feed-forward network $\mathcal{N}$ with parameters $\theta \in \mathbb{R}^{M}$ and a sigmoid activation $sig(\cdot)$ . We adopt mean-square error (MSE) as the distillation loss to transfer capacity:
\begin{align}
    \mathcal{L}_{KD} = (sig(\theta_s x_s + \theta_d x_d) - sig(\theta_s x_s))^2
\end{align}
where $\theta_s \subseteq \theta$ is the chosen parameters, corresponding to the subnets in the manuscript; $\theta_d = \theta \setminus \theta_s$ is the unchosen parameters; $x_s$ and $x_d$ are the inputs cooperating with $\theta_s$ and $\theta_d$, respectively. Note that, to avoid teacher degradation in vanilla KD, the teacher supervision, i.e., $sig(\theta_s x_s + \theta_d x_d)$, is detached from the computation graph, thus irrelevant to computing the partial gradient of $\theta_s$: 
\begin{align}
\label{eq:partial}
   \frac{\partial \mathcal{L}_{KD}}{\partial \theta_s} = -2[sig(\theta_sx_s+\theta_dx_d) - sig(\theta_sx_s)] \cdot sig(\theta_sx_s) \cdot [1-sig(\theta_sx_s)] \cdot x_s.
\end{align}
This partial gradient consists of three components: 1) $sig(\theta_sx_s+\theta_dx_d) - sig(\theta_sx_s)$, 2) $sig(\theta_sx_s) \cdot [1-sig(\theta_sx_s)]$, and 3) $x_s$. When the optimization converges, at least one component is deemed to approach zero. Since $x_s$ is the input following the distribution of data, it is not guaranteed to approach zero. Similar to~\cite{ye2022stimulative,ye2023stimulative}, we apply Taylor expansion to analyze $[sig(\theta_sx_s+\theta_dx_d) - sig(\theta_sx_s)]$:
\begin{align}
\label{eq:taylor}
    sig(\theta_sx_s+\theta_dx_d) - sig(\theta_sx_s) \approx sig(\theta_sx_s) \cdot [1-sig(\theta_sx_s)] \cdot \theta_dx_d.
\end{align}
Note that $\theta_dx_d$ can be considered as a constant from two aspects: 1) $\theta_d$ keeps unaltered due to the absence of its gradient flow; 2) $x_d$ is the input data and invariant to optimization. Consequently, the partial gradient $\partial \mathcal{L}_{KD}/\partial \theta_s$ is approximately determined by $sig(\theta_sx_s)\cdot[1-sig(\theta_sx_s)]$. If the $L_1$ norm of $\theta_s$ increases dramatically, it is likely to make the partial gradient approach zero. To sum up, optimizing $\mathcal{L}_{KD}$ produces no influence on $\theta_d$ and enhances the $L_1$ norm of $\theta_s$, which results in the relative sparsity effect. Since qualitative analyses and approximations are involved in the above mathematical proof, rigorousness is not guaranteed. This section merely provides an illustrative understanding of the relative sparsity effect in ST. Delving into rigorous proofs in more complex situations could become a future research direction, but this is beyond the scope of discussion in this manuscript.

\section*{Appendix F: Efficiency of STP}
To explore the efficiency of STP, we record the training time of STP and other pruning methods. The experiments are conducted on CIFAR100, using ResNet50 as the backbone. As shown in Table~\ref{tab:training time}, our method achieves superior performance with acceptable (the second shortest) training time. The efficiency of our method mainly lies in high parallelism. The forward/backward passes of the main network and subnets can be executed at the same time leveraging CUDA streams or multi-GPU model parallelism. Besides, our method is one-stage, eliminating sequential fine-tuning or iterative pruning, which can further reduce the training time. To ablate the influence of forward/backward counts, we extend the epochs of other methods three times to align our forward/backward counts. With 3x forward/backward, other methods struggle to improve further. It demonstrates that merely extending the training schedule cannot notably enhance pruning performance, suggesting the superiority of our method.
\begin{table}[ht]
\caption{Performance and training time of different methods and settings.}
\resizebox{\linewidth}{!}{
\tiny
\begin{tabular}{c|cccc|c}
\hline
                      & RST-S & Depgraph & OTO v2 & IMP-Refill & Ours   \\ \hline
Top-1 Acc(1x training schedule)  & 75.02  & 49.07    & 77.04  & 75.12      & 79.64  \\
Top-1 Acc(3x training schedule) & 75.39  & 50.12    & 77.34  & 75.97      & -      \\
GPU time per epoch    & 46.58s & 75.10s   & 82.22s & 77.32s     & 61.92s \\ \hline
\end{tabular}
}
\label{tab:training time}
\end{table}

\end{document}